\def\year{2021}\relax
\documentclass[letterpaper]{article} 
\usepackage{aaai21}  
\usepackage{times}  
\usepackage{helvet} 
\usepackage{courier}  
\usepackage[hyphens]{url}  
\usepackage{graphicx} 
\usepackage{natbib}  
\usepackage{caption} 
\usepackage{amsfonts}
\usepackage{amsmath}
\usepackage{wrapfig}
\usepackage[switch]{lineno}

\usepackage[boxed,ruled,linesnumbered]{algorithm2e} 

\SetCommentSty{mycommfont}
\let\oldnl\nl

\SetAlCapHSkip{0pt}
\IncMargin{-\parindent}
\usepackage{color}
\definecolor{green}{rgb}{0, 0.5, 0}
\definecolor{orange}{rgb}{0.8, 0.6, 0.2}
\definecolor{red}{rgb}{1.0, 0.0, 0.0}
\definecolor{teal}{rgb}{0.0, 0.4, 0.4}
\definecolor{purple}{rgb}{0.65,0,0.65}
\definecolor{saffron}{rgb}{0.95,0.75,0.2}
\definecolor{turquoise}{rgb}{0.0,0.5,0.5}
\newcommand{\nonl}{\renewcommand{\nl}{\let\nl\oldnl}}
\newlength\savedwidth

\title{Second-order Neural Network Training Using Complex-step Directional Derivative}

\author {
    Siyuan Shen,\textsuperscript{\rm 1}
    Tianjia Shao, \textsuperscript{\rm 1}
    Kun Zhou, \textsuperscript{\rm 1}
    Chenfanfu Jiang,\textsuperscript{\rm 2}
    Feng Luo, \textsuperscript{\rm 3}
    Yin Yang \textsuperscript{\rm 3}\thanks{Corresponding author.}
    \\
}
\affiliations {
    \textsuperscript{\rm 1}
    State Key Lab of CAD\&CG Zhejiang University \\
    \textsuperscript{\rm 2} Penn Institute for Computational Science, University of Pennsylvania \\
    \textsuperscript{\rm 3} School of Computing, Clemson University \\
    \{shensiyuan,tjshao\}@zju.edu.cn,
    kunzhou@acm.org,
    cffjiang@seas.upenn.edu,
    \{luofeng,yin5\}@clemson.edu
}

\begin{document}
\maketitle

\begin{abstract}
While the superior performance of second-order optimization methods such as Newton's method is well known, they are hardly used in practice for deep learning because neither assembling the Hessian matrix nor calculating its inverse is feasible for large-scale problems. Existing second-order methods resort to various diagonal or low-rank approximations of the Hessian, which often fail to capture necessary curvature information to generate a substantial improvement. On the other hand, when training becomes batch-based (i.e., stochastic), noisy second-order information easily contaminates the training procedure unless expensive safeguard is employed. As a result, first-order methods prevail and remain the dominant solution for modern deep architectures. In this paper, we adopt a numerical algorithm for second-order neural network training. We tackle the practical obstacle of Hessian calculation by using the complex-step finite difference (CSFD) -- a numerical procedure adding an imaginary perturbation to the function for derivative computation. CSFD is highly robust, efficient, and accurate (as accurate as the analytic result). This method allows us to literally apply any known second-order optimization methods for deep learning training. Based on it, we design an effective Newton Krylov procedure. The key mechanism is to terminate the stochastic Krylov iteration as soon as a disturbing direction is found so that unnecessary computation can be avoided. During the optimization, we monitor the approximation error in the Taylor expansion to adjust the step size. This strategy combines advantages of line search and trust region methods making our method preserves good local and global convergency at the same time. We have tested our methods in various deep learning tasks. The experiments show that our method outperforms exiting methods, and it often converges one-order faster. We believe our method will inspire a wide-range of new algorithms for deep learning and numerical optimization.
\end{abstract}

\section{Introduction}\label{sec:intro}
Given the training set, we consider a neural network a function $f(w)$ which takes $w \in \mathbb{R}^{N}$, the vector of network parameters as the input and outputs a real loss value. The training starts with an initial guess of $w_0$ aiming to improve $w$ over iterations $w_0, w_1,\cdots$ until $w_k$ is sufficiently close to the minimizer $w^*$. At the $k$-th iteration an improvement of $\Delta w_k$ is calculated, which updates $w$ by $w_{k+1} = w_k + \Delta w_k$. One could Taylor expand $f(w_{k+1})$ as:
\begin{equation}\label{eq:taylor}
\begin{split}
f(w_{k+1}) & = f(w_k + \Delta w_k) =  f(w_k) + \nabla f(w_k) \cdot \Delta w_k \\
& + \frac{1}{2} \Delta w_k^\top \nabla^2 f(w_k) \Delta w_k + \mathbf{O}(\|\Delta w_k \|^3).
\end{split}
\end{equation}
If we ignore the high-order error term of $\mathbf{O}(\|\Delta w_k \|^3$, $\Delta w_k$ can be computed as $\Delta w_k = -H_k^{-1} g_k$, where $g_k = \nabla f(w_k)$ and $H_k = \nabla^2 f(w_k)$ are the gradient and Hessian of $f(w_k)$. Eq.~\eqref{eq:taylor} lays the foundation of the Newton's method, which is probably one of the most powerful optimization methods we are aware of, converging at a quadratic rate locally.

Despite the strong desire of harvesting the power of quadratic convergency, Newton's method is much less used in practice as its advantages are overshadowed by several algorithmic and practical obstacles. First, it is difficult to analytically calculate $H$ for deep networks. Computing $H$ involves a tedious second-order differentiation chain along the net. While some auto differentiation (AD) techniques~\cite{paszke2017automatic} offer derivative calculations, they often become clumsy for second-order derivatives making the implementation costly, labor-intensive and error-prone. $H$ is a high-dimension \emph{dense} matrix (i.e., $ H \in \mathbb{R}^{N \times N}$). Thus, it should never be explicitly assembled, ruling out all the direct linear solvers like LU or Cholesky. Newton's method is also numerically sensitive: an ill-conditioned Hessian yields ``dangerous'' $\Delta w$ and crashes the optimization quickly. Bigger training sets further exacerbate those challenges as we are actually performing online or \emph{stochastic} optimization by sampling the actual Hessian at each batch, which could be noisy and misleading. Due to these reasons, even we have witnessed several elegant pseudo second-order techniques like SMD~\cite{schraudolph2002fast,martens2010deep},  AdaGrad~\cite{duchi2011adaptive}, and Shampoo~\cite{gupta2018shampoo,anil2020second} etc, many of them remain gradient-based or quasi-Newton-like. The information hidden in $H$ is seldom fully exploited.

In this paper, we propose an algorithmic solution that leverages (stochastic) Newton to train deep neural nets using numerically computed Hessian. More precisely, we compute the first- and second-order numerical directional derivative of $\nabla f$ and $f$, which give Hessian-vector products ($Hp$) and Hessian-inner product ($p^\top H p$) for a vector $p$. While this fact has been used previously for machine learning~\cite{schraudolph2002fast}, we offer a new implementation of this idea with very little extra coding work under commonly-used deep learning frameworks like \texttt{TensorFlow}~\cite{abadi2016tensorflow} and \texttt{PyTorch}~\cite{paszke2019pytorch}. We name our method \emph{complex-step directional derivative} or CSDD. Compared with the classic finite difference method (FD)~\cite{forsythe1960finite}, CSDD is highly robust, accurate, and can be easily generalized to higher-order cases. CSDD relieves implementation barriers of second-order training and seamlessly couples with a wide-range of Newton and quasi-Newton based methods~\cite{knoll2004jacobian}. Based on CSDD, we advocate a novel stochastic Newton Krylov method for second-order network training. Our method integrates advantages of both line search and trust region methods and fully leverages the curvature information whenever it is reliable. Instead of setting a region radius, we watch the Taylor expansion error and early terminate the Krylov iteration as soon as a risky direction is identified. This approach is fundamentally different from other alternatives like least-square CG (LSCG) or Levenberg–Marquardt (LM) method as we do not alter the shape of the optimization manifold. Therefore, every iteration is descent leading to a better loss. We have tested our method in various deep learning scenarios, and our method consistently outperforms existing competitors. We observe strong second-order convergency in many situations, often one-order faster than commonly-used methods like Adam or Shampoo.

\section{Related Work}\label{related}
The prosperity of deep learning architectures and their applications is not possible without the nutrition from underneath optimization and numerical methods. The training procedure in deep learning is normally regarded as a nonlinear optimization, which relies on the information of gradient and/or Hessian of the target function. On the other hand, modern deep networks are often too complex to be analytically formulated, which can only be dealt with differentiable algorithms like backpropagation~\cite{rumelhart1986learning} also known as BP. BP is a special implementation of AD~\cite{al2008nonlinear,paszke2017automatic}. It computes the gradient of the loss function via the chain rule, layer by layer along the network. Based on the it, first-order methods such as SGD~\cite{bottou2010large}, Adam~\cite{dozat2016incorporating,kingma2014adam}, AdaGrad~\cite{duchi2011adaptive}, RMSprop~\cite{schaul2013no,sutskever2013importance}, etc. flourish with increased performance and robustness.

The theories of second-order methods have been well studied ~\cite{nocedal2006numerical,ueberhuber2012numerical}. Generalizing BP to retrieve second-order information as in~\cite{becker1988improving,mizutani2008second} seems quite possible for deep learning at first sight. Yet, its actual deployment is less common than first-order methods. This is probably because AD techniques become cumbersome in higher-order generalization. Overloading the analytic second-order differentiation is significantly more involved than the first-order case~\cite{margossian2019review}, and performing first-order differentiation multiple times to obtain a high-order derivative could lead to inefficient and numerically unstable code~\cite{betancourt2018geometric}. For deep learning, Hessian is a dense matrix, and it may not even fit into the memory. Therefore, many research efforts investigate the possibility of simplifying Hessian using for instance, diagonal~\cite{chapelle2011improved}, diagonal block~\cite{botev2017practical,martens2015optimizing,zhang2018local}, and low-rank~\cite{anil2020second,gupta2018shampoo} approximations. While they are able to circumvent the memory issue, such simplification does not expose the full spectrum of the curvature, and we can barely observe quadratic convergency in practice. We would also like to point out that the phrase ``Hessian'' in existing literature could be misleading. Some previous contributions regarded the network as a nested function: $f(w)=l\left(P(w)\right)$, where $l(\cdot)$ is the loss function mapping a network prediction $P(w)$ to the final loss value. Instead of computing $H = \nabla^2 f$, the Jacobian of $P$, $J=\partial P/\partial w$, was used as a Gauss-Newton approximation of $H$~\cite{martens2010deep,schraudolph2002fast,amari2000adaptive}. $J^\top J$ does partially constitute $H$, and one can quickly verify this fact by examining the second-order chain rule of $f(w)=l\left(P(w)\right)$. Yet, Gauss-Newton method remains a first-other procedure.

On the other hand, Hessian-free (HF) method seems to be a more attractive option. Since $H$ should not be explicitly built, HF only calculates projected $H$ i.e., the Hessian-vector product, which suffices for many non-direct solvers like Newton Krylov methods~\cite{knoll2004jacobian}. Calculating Hessian-vector product is equivalent to calculating the \emph{directional derivative} of the gradient. There are several choices out there. We can use symbolic differentiation (SD) method like the $\mathcal{R}\{\cdot\}$ technique~\cite{pearlmutter1994fast}, a numerical derivative like FD, or AD-based solutions. Unfortunately, none of them offers both efficiency and accuracy. FD is the most efficient, but subject to numerical issues. SD and AD are accurate, which essentially compute the analytic differentiation via the chain rule. However, they suffer a high overhead computation because of axillary data structures used (e.g., the computation graph etc.).

Alternatively, we use CSDD to facilitate HF computation. As to be detailed in the next section, CSDD is robust, accurate, and more efficient than existing AD packages e.g., CSDD is $14\%$ faster than \texttt{Tensorflow} and $27\%$ faster than \texttt{PyTorch}. Based on CSDD, we re-examine the classic Newton-based optimization techniques and bring several non-trivial enhancements/improvements. During the training, we fully leverage the global gradient to sift noisy batches, and investigate computation efforts only to ``worthy'' batches. We do not try to globally alter pathological curvature as in LSCG or LM. Instead, we identify risky regions and skip unnecessary Hessian-related computation as much as possible. To the best of our knowledge, this method is the first attempt to synergize CSFD with deep learning and offer a full second-order solution. Our experiments demonstrate a strong convergency behavior on various deep learning tasks.

\section{Complex-step Finite Difference}\label{sec:CSFD}
In order to make the paper more self-contained, we start our discussion with a brief review of the finite difference method, its numerical issue of the subtractive cancellation, and its generalization with complex arithmetic.

\subsection{Subtractive cancellation of finite difference}
Given a function $f:\mathbb{R}\rightarrow\mathbb{R}$ differentiable around $x=x_0$. Under a small perturbation $h$, $f$ can be first-order Taylor expanded as $f(x_0 + h) = f(x_0) + f'(x_0) \cdot h + \mathbf{O}(h^2)$, leading to the forward finite difference (FFD) scheme:
\begin{equation}\label{eq:fd}
 f'(x_0) \approx \frac{f(x_0 + h) - f(x_0)}{h}.
\end{equation}
Eq.~\eqref{eq:fd} suggests that $h$ should be as small as possible for a good approximation. Unfortunately, as a computer has limited bits to digitalize  real numbers, all the floating-point arithmetics have the round-off error~\cite{ueberhuber2012numerical}, a small relative error also known as \emph{machine epsilon} $\epsilon$. For the double precision of \texttt{IEEE~754} floating-point standard~\cite{ieee1985ieee}, $\epsilon\approx 1.11 \times 10^{-16}$. Normally, the round-off error does not seriously impair the stability or the accuracy of a numerical procedure. However, when $h$ gets smaller, $f(x_0+h)$ and $f(x_0)$ become nearly equal to each other. Subtraction between them would eliminate \emph{leading} significant digits, and the result after the rounding could largely deviate from the actual value of $f(x_0 + h)-f(x_0)$.
%
%
%
Some numerical literature e.g., \cite{nocedal2006numerical} considers that central finite difference (CFD) with the form:
\begin{equation}\label{eq:cd}
f'(x_0)\approx\frac{f(x_0+h)-f(x_0-h)}{2h},
\end{equation}
has higher accuracy, with a quadratic error term of $\mathbf{O}(h^2)$. This conclusion is only licit when the subtractive cancellation does not occur. In reality, CFD could be even more sensitive to a smaller $h$ because of its faster convergent rate.

\subsection{First-order complex-step finite difference}
\begin{figure}[ht!]
  \centering
  \includegraphics[width=0.95\linewidth]{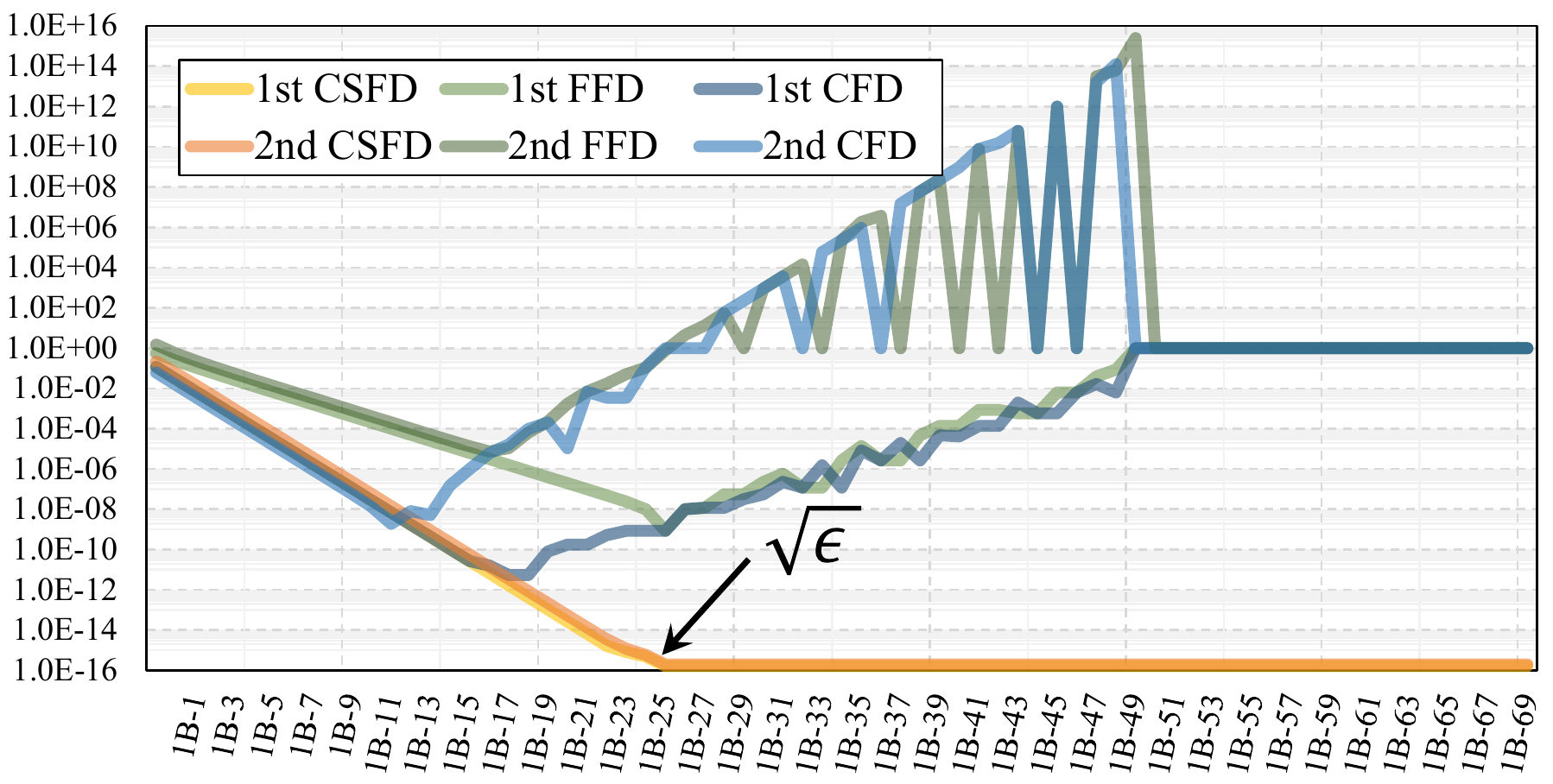}
  \caption{Relative error of different numerical differentiation schemes of $f(x)=e^x / (x^2 + 1)$ at $x = 10$.}\label{fig:CSFD_error}
\end{figure}

The subtractive cancellation can be avoided by using the complex-step finite difference (CSFD)~\cite{martins2003complex}, which is based on the complex Taylor series expansion~\cite{lyness1968differentiation}:
\begin{equation}\label{eq:complex_taylor}
f^*(x_0 + hi) = f^*(x_0) + f^{*'}(x_0)\cdot{hi} +\mathbf{O}(h^2).
\end{equation}
Here, we promote a real-value function $f$ to be a complex-value one $f^*$ by allowing complex inputs while retaining its original computation procedure. Under this circumstance, we have $f^*(x_0) = f(x_0) \in \mathbb{R}$ and $f^{*'}(x_0) = f'(x_0) \in \mathbb{R}$. Extracting imaginary parts of both sides in Eq.~\eqref{eq:complex_taylor} yields $\mathtt{Im}\big(f^*(x_0 + hi)\big) = \mathtt{Im}\big(f^*(x_0)+f^{*'}(x_0)\cdot{hi}\big)+\mathbf{O}(h^3)$. We then have the first-order CSFD formulation:
\begin{equation}\label{eq:cfd}
f'(x_0)=\frac{\mathtt{Im}\big(f^*(x_0+hi)\big)}{h}+\mathbf{O}(h^2)\approx\frac{\mathtt{Im}\big(f^*(x_0+hi)\big)}{h}.
\end{equation}
It is clear that Eq.~\eqref{eq:cfd} does not have a subtractive numerator meaning it only has the round-off error regardless of the size of the perturbation $h$. In addition, the operation of $\mathtt{Im}(\cdot)$ removes the $(hi)^2$ term in the complex Taylor expansion reducing the approximation error to $\mathbf{O}(h^2)$. If $h \sim \sqrt{\epsilon}$ i.e., around $2\times10^{-24}$ in Fig.~\ref{fig:CSFD_error}, CSFD approximation error is at the order of $\epsilon$. Hence, CSFD can be as accurate as analytic derivative as analytic derivative also has round-off of $\epsilon$.

An example is plotted in Fig.~\ref{fig:CSFD_error}, where we compare the relative error of numerical derivatives of $f(x) = e^x / (x^2 + 1)$ using FFD, CFD, and CSFD with its analytic derivative at $x = 10$. The numerical behavior of FFD and CFD is consistent with our analysis: when the perturbation $h$ decreases, CFD converges faster than FFD initially. Both FFD and CFD soon hit the threshold of subtractive cancellation. After that, the relative error bounces back immediately. CSFD reduces the error as quickly as CFD, and the relative error stably remains at the order of $\epsilon$.

\subsection{Second-order complex-step finite difference}
The generalization of CSFD to second- or even higher-order differentiation is straightforward by making the perturbation a \emph{multicomplex} quantity~\cite{lantoine2012using,nasir2013new}. The multicomplex number is defined recursively: its base cases are the real set $\mathbb{C}^0 = \mathbb{R}$, and the regular complex set $\mathbb{C}^1 = \mathbb{C}$. $\mathbb{C}^1$ extends the real set ($\mathbb{C}^0$) by adding an imaginary unit $i$ as: $\mathbb{C}^1 = \{x + yi |x,y \in \mathbb{C}^0 \}$. The multicomplex number up to an order of $n$ is defined as: $\mathbb{C}^n = \{ z_1 + z_2 i_n | z_1, z_2 \in \mathbb{C}^{n-1} \}$. Under this generalization, the multicomplex Taylor expansion becomes:
\begin{equation}\label{eq:mc_taylor}
\begin{split}
&f^{\star}(x_0+hi_1+\cdots+hi_n)=f^{\star}(x_0)+f^{\star'}(x_0) h\sum_{j=1}^ni_j \\
&+\frac{f^{\star''}(x_0)}{2} h^2 \big(\sum_{j=1}^ni_j\big)^2
+\cdots+\frac{f^{\star(k)}}{k!} h^k \big(\sum_{j=1}^ni_j\big)^k\cdots.
\end{split}
\end{equation}
Here, $\left(\sum i_j\right)^k$ can be computed following the \emph{multinomial theorem}, and it contains products of mixed $k$ imaginary directions for $k$-th-order terms. The second-order CSFD formulation can then be derived as follows:
\begin{equation}\label{eq:mcfd_hessian}
\begin{array}{l}
  \displaystyle \frac{\partial^2f(x,y)}{\partial x^2} \approx \frac{\mathtt{Im}^{(2)}\big(f(x+hi_1+hi_2, y)\big)}{h^2},\\


  \displaystyle \frac{\partial^2f(x,y)}{\partial x \partial y} = \frac{\partial^2f(x,y)}{\partial y \partial x} \approx \frac{\mathtt{Im}^{(2)}\big(f(x+hi_1, y+hi_2)\big)}{h^2},
\end{array}
\end{equation}
where $\mathtt{Im}^{(2)}$ picks the mixed imaginary direction of $i_1 i_2$. One can easily tell from Eq.~\eqref{eq:mcfd_hessian} that second-order CSFD is also subtraction-free making them as robust/accurate as the first-order case (e.g., see Fig.~\ref{fig:CSFD_error}). Its recursive definition also greatly eases the implementation.

\section{Complex-step Directional Derivative}\label{sec:CSDD}
It may look pointless to have CSDD as the ordinary CSFD alone is able to compute gradient and Hessian accurately. As we will see in this section, CSDD allows us to \emph{collectively} apply the perturbation along $p$ instead of perturbing every element in $w$. Therefore, CSDD better collaborates with existing deep learning frameworks rather than using CSFD for all the differential operations.

As the name implies, CSDD uses CSFD to calculate projected Hessian under a given direction $p$. For instance, $Hp$ can be written as:
\begin{equation}\label{eq:Hp1}
[H(w)p]_a = \sum_{b=0}^{N-1} \lim_{h \rightarrow 0}\frac{[g(w + h e_b) - g(w)]_a}{h} \cdot [p]_b,
\end{equation}
where $[v]_a \in \mathbb{R}$ returns $a$-th element in vector $v$; $e_b \in \mathbb{R}^N$ is a vector with all the elements being zero expect for the $b$-th one, which equals to one. When $h \rightarrow 0$, $[p]_b h \rightarrow 0$, and we substitute $h$ with $[p]_b h$ to cancel the multiplication of $[p]_b$:
\begin{equation}\label{eq:Hp2}
[H(w)p]_a = \sum_{b=0}^{N-1} \lim_{h \rightarrow 0}\frac{[g(w +  [p]_b h  e_b) - g(w)]_a}{h}.
\end{equation}
Following the linearity of directional derivative, we obtain:
\begin{equation}\label{eq:Hp}
Hp = \lim_{h \rightarrow 0} \frac{g(w + h p) - g(w)}{h} \approx \frac{\mathtt{Im}(g(w + hi \cdot p))}{h}.
\end{equation}
Similarly, the quadratic form of $p^\top H p$ is the second-order directional derivative $\nabla^2_p f$. Therefore, we have:
\begin{equation}\label{eq:pHp}
\begin{split}
p^\top Hp & = \lim_{h_1, h_2 \rightarrow 0} \frac{h_1 p^\top \big (\nabla f(w + h_2 p ) - \nabla f(w)\big)}{h_1 h_2}\\
&\approx \frac{\mathtt{Im}^{(2)}(f (w + (hi_1 + hi_2) \cdot p))}{h^2}.
\end{split}
\end{equation}
From Eqs.~\eqref{eq:Hp} and~\eqref{eq:pHp}, it is noticed that we only need to apply a single perturbation to compute $Hp$ or $p^\top H p$. This computation can be done with one complex-enabled forward and backward pass of the network. On the other hand, computing $g$ or $H$ requires $N$ and $N^2$ perturbations, not to mention the memory consumption. In theory, the CSFD is more efficient than BP or other AD-based subroutines if properly implemented. However, we found CSDD a more feasible option for the purpose of maximizing the re-usability of existing deep learning code. With CSDD, we extract necessary second-order information to devise a full second-order training algorithm, which is to be discussed in the next section.

\section{Stochastic Newton Krylov Optimization}\label{sec:newton}
It is known that conjugate gradient (CG) is highly effective for SPD systems with clustered spectra. CG minimizes errors within a Krylov subspace that is iteratively spanned. During the iteration, CG only computes $p^\top Hp$ and $Hp$ for a search direction $p$, making itself an ideal HF candidate to solve the Newton step of $H \Delta w = -g$~\cite{martens2010deep,chapelle2011improved,vinyals2012krylov,martens2011learning}. Here, we discard the subscript $(\cdot)_k$ of the Newton iteration index for a more concise notation.

\subsection{Challenges in stochastic optimization}
When $H$ has vanished eigenvalues, CG may experience the division-by-zero issue, and it also diverges if $H$ has negative eigenvalues. In practice, we only know $H$ is symmetric, and there is no guarantee for its positive definiteness. A commonly used strategy is to employ conjugate residual~\cite{saad2003iterative} or LSCG~\cite{toh2002solving}, which solve a least-square Newton step of $H^\top H \Delta w = -H^\top g$. The singularity of the system, on the other hand, is overcome by using LM method~\cite{more1978levenberg} -- adding a small diagonal damping to restore the positive definiteness of $H$ or $H^\top H$.
\begin{figure}[ht!]
  \centering
  \includegraphics[width=0.99\linewidth]{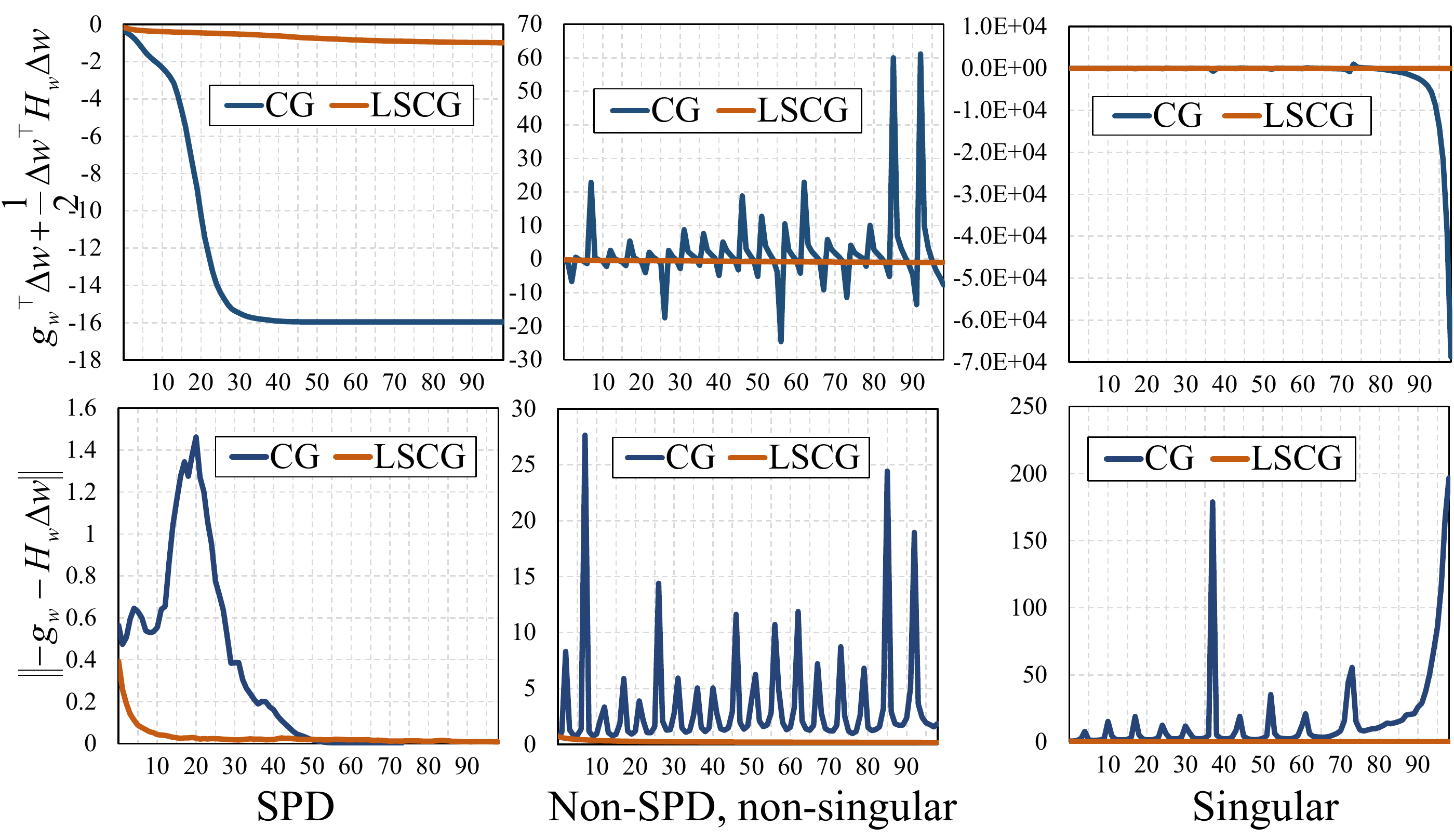}
  \caption{CG is designed to reduce the quadratic energy of $E_Q = g^\top \Delta w + \frac{1}{2}\Delta w^\top H \Delta w$. Focusing on the residual of the linear system could be misleading. LSCG, on the other hand, covers the curvature information and should only be used with extra cautions in stochastic optimization.}\label{fig:LSCG}
\end{figure}

Unfortunately, none of these methods offers a true remedy of Newton-CG for deep learning. In stochastic optimization, \emph{non positive definiteness suggests either problematic optimization regions or noises in batch-based sampling}. We ought to invest computing efforts in neither case. The reason is simple: if the local curvature is misshaped, a complete Newton step that fully solves $H \Delta w = -g$ is unlikely profitable. By projecting this linear system into the column space of $H$, LSCG measures a transformed residual error, which ignores the sign of the curvature; LM evenly ``bends'' the curvature of the Hessian so that local flatness becomes convex. Clearly, the original curvature information is lost in LSCG and LM. If $w$ is distant from $w^*$, such Hessian modifications could adversely affect the global convergency.

Another important observation is that CG is actually an algorithm for minimizing a quadratic energy of:
\begin{equation}\label{eq:quadratic_error}
E_{Q} = \Delta w^\top g + \frac{1}{2}\Delta w^\top H \Delta w,
\end{equation}
which is a different measure from the residual error ($\|r\|$). As we can see from the left column of Fig.~\ref{fig:LSCG}, even with a SPD system, residual error can still go up during CG iterations, while $E_{Q}$ monotonically decreases. When $H$ is singular and semi-SPD, CG iterations also reduce $E_{Q}$ unless a division-by-zero error occurs. However, if $H$ is non-SPD with both positive and negative eigenvalues, CG iterations are in limbo with an oscillating $E_Q$ and could diverge any time. Interestingly, it is noted that $E_Q$ also appears in the Taylor expansion of $f(w_{k+1})$ i.e., Eq.~\eqref{eq:taylor}. Since $f(w_k)$ is a fixed quantity, minimizing $f(w_{k+1})$ is equivalent to minimizing $E_Q + \mathbf{O}\|\Delta w\|^3$. This implies that \emph{if $\mathbf{O}\|\Delta w\|^3$ is not the dominating factor in Eq.~\eqref{eq:taylor}, a CG iteration should be descending under a non-negative curvature}.

\subsection{Our method}
Bearing above analysis and observations in mind, we propose an improved stochastic Newton Krylov procedure. Our method is empowered by first- and second-order CSDD (i.e., $\mathtt{CSDD}$ and $\mathtt{CSDD2}$ in Alg.~\ref{alg:our}). We do not rely on Hessian modifications as in LSCG or LM. Instead, we exploit $\mathtt{CSDD2}$ to switch among different search strategies. Our method significantly improves the convergency performance and is not sensitive to parameter tuning. As one can see from Alg.~\ref{alg:our}, our algorithm consists of three major subroutines (delimited by dotted lines), which are to be detailed as follows.

\setlength{\textfloatsep}{15 pt}
\begin{algorithm}[t!]
\footnotesize
\SetNlSty{small}{}{:}

\KwIn{minibatch set $\{\mathcal{B}_1, \mathcal{B}_2, \cdots \}$, $\widetilde{\eta}$}
\vspace{-5 pt}
\nonl\hrulefill\\
compute $g$; \tcp{$g$ is the global gradient}
$j \leftarrow 0$; \tcp{minibatch index}
\For {each $\mathcal{B}_j$}
{
    compute $g_j$; \tcp{$g_j$ is the local gradient}
    \If {$g_j \cdot g < 0$}
    {
        $j \leftarrow j + 1$;\\
        continue; \tcp{skip this loop for $\mathcal{B}_j$}
    }
    \nonl\dotfill \\
    $i \leftarrow 0$, $\quad \Delta w \leftarrow \Delta \widetilde{w}$;\\
    $ \langle g, h\rangle \leftarrow \mathtt{CSDD}(w, \Delta w) $;\\
    $ r_i \leftarrow -(g + h_i)$, $\quad p_i \leftarrow r_j$;\\
   \While {$\| r_i \|$ is not small enough}
    {
        $\kappa \leftarrow \mathtt{CSDD2}(w, p_i)$; \tcp{$\kappa = p_i^\top H p_i$}
        \If {$\kappa < 0$}
        {
            break; \tcp{early termination}
        }
        \If {$0 < \kappa < 1.0 \times 10^{-8}$}
        {
            $\kappa \leftarrow 0.01\cdot \|p_j\|$;
        }
        $\alpha_i \leftarrow \| r_i\| / \kappa$;\\
        $\Delta w \leftarrow \Delta w + \alpha_i p_i$;\\
        $ q_i \leftarrow \mathtt{CSDD}(w, p_i)$;\\
        $r_{i+1} \leftarrow r_i - \alpha_i q_i$; \\
        $\beta_j \leftarrow \| r_{i+1} \|^2 / \| r_j \|^2$;\\
        $\displaystyle p_{i+1} \leftarrow r_{i+1} + \beta_i p _i$;\\
        $\displaystyle i \leftarrow i + 1$;
    }
    \If {$\Delta w \cdot g > 0$}
    {
        $\Delta w \leftarrow \big(1+\frac{\Delta w \cdot g}{\|\Delta w\| \cdot \|g\|}\big)\Delta w$;\\
    }
    \nonl\dotfill \\
    $\gamma \leftarrow 1$; \tcp{the default step size}
    compute $\eta$;\\
    \If {$\eta > 0.5\widetilde{\eta}$}
    {
        $s \leftarrow |\widetilde{\eta} / \eta|$; \tcp{an initial test size}
        \While{$\eta \notin [0.5\widetilde{\eta}, \widetilde{\eta}] $}
        {
            \If{$\eta > \widetilde{\eta}$}
            {
                $\gamma \leftarrow 0.5 \cdot \gamma$; \tcp{shrink a step}
            }
            \Else
            {
                $\gamma \leftarrow 1.5 \cdot \gamma$; \tcp{stretch a step}
            }
            update $\eta$ with $f(w + \gamma \cdot \Delta w)$;
        }
    }
    $w \leftarrow w + \gamma \cdot \Delta w$;\\
    compute $g$;\\
    $\displaystyle j \leftarrow j + 1$;
}
\caption{Our stochastic Newton CG method.}
\label{alg:our}
\end{algorithm}

\paragraph{Pre- and post- batch screening}
We examine if a minibatch is potentially noisy or unreliable. If yes, we simply skip it to avoid any Hessian computations for this batch. A so-called unreliable batch is defined based on the consistence between local and global gradients: if the batch gradient ($g_i$) is opposite to the global gradient ($g$) i.e., $g_i \cdot g < 0$, the minibatch $\mathcal{B}_i$ is discarded.

The post-batch screening occurs after we finish needed Krylov iterations at a local batch. We examine if $\Delta w$ calculated is consistent with negative global gradient $-g$. In theory, Krylov subspace is spanned by conjugating the current residual, and $-g \cdot \Delta w$ is always non-negative. In practice however, as $H$ is sub-sampled, even with pre-batch screening, $\Delta w$ occasionally deviates from $-g$. The post-batch screening removes any search components along $g$  (line 29) so that $\Delta w$ does not cancel previous improvements and remains a global descent.

\paragraph{Krylov loop with early termination}
When a minibatch $\mathcal{B}_i$ is considered reliable, the algorithm steps into a CG-like Krylov iteration (lines 12 -- 27). At each iteration, we monitor the second-order directional derivative of the local loss $\nabla^2_p f_i$ under the search direction $p$. $\nabla^2_p f_i$ reveals the local curvature along $p$, and we terminate the Krylov loop for $\mathcal{B}_i$ as soon as $\nabla^2_p f_i$ becomes negative (lines 14 -- 16). This negativeness may not reflect the true configuration of the global Hessian but a miss-representation induced by sub-sampling. Therefore, we do not quit the Newton step immediately, but switch to another Hessian sample at the next (reliable) batch. This strategy can also be understood as splitting the full set of global CG iterations over multiple local Hessian samples, and it becomes a classic CG when a local Hessian is highly representative and positive definite. However, it never happens in our experiments, and a local Krylov procedure seldom iterates more than 20 loops.

On the other hand, if $\nabla^2_p f_i$ is close to zero suggesting some local linearity. As discussed before (i.e., see Fig.~\ref{fig:LSCG}), standard CG still improves $E_Q$ in this situation. Therefore, we apply a \emph{temporary} momentum of $0.01\cdot \|p_j\|$ (line 18) to avoid the division-by-zero problem and push the search out of the flat zone. This numerical treatment is different from LM. LM \emph{globally} and \emph{permanently} modifies the geometry of the Hessian, while our algorithm alters the curvature locally and temporarily. In addition, LM uses a fixed global damper, which ideally should be just enough to compensate the smallest negative eigenvalue of $H$. Unfortunately, as we are agnostic to $H$ and its spectrum, finding a proper damping is tricky or sometimes even impossible. Our method is adaptive: the local momentum is set based on current search velocity ($0.01\cdot\|p_i\|$) -- it works robustly by default.

\paragraph{Error-based step sizing}
The goal of reducing $f(w_{k+1})$ can be achieved by minimizing $E_Q$ only when the quadratic Taylor approximation of $f(w_{k+1})$ is appropriate. In other words, we would like to keep $\mathbf{O}\|\Delta w\|^3$ reasonably small during the optimization. This idea has led to several variations of Newton's method with \emph{cubic regularization}~\cite{song2019inexact,benson2018cubic}, and has been tested in learning tasks~\cite{kovalev2019stochastic}. We choose a similar but more effective approach with the help of CSDD by directly calculating the \emph{Taylor ratio}, the ratio between the second-order Taylor approximation error and loss reduction:
\begin{equation}\label{eq:taylor_err}
\eta = |\frac{f(w+\Delta w) - f(w) - E_Q}{f(w+\Delta w) - f(w)}|.
\end{equation}
Since we only search under positive local curvatures, $\|\Delta w\|$ increases monotonically. Along the iteration of reducing $E_Q$, the second-order Taylor approximation also becomes less representative, and $\eta$ accurately measures this trade-off.
The global convergency of Newton method is normally implemented by adjusting the step size based on Wolfe conditions~\cite{wolfe1969convergence,wolfe1971convergence}. However, we found that $\eta$ is a more effective tool for this purpose. Specifically, we stretch or shrink the step size $\gamma$ by making sure $\eta$ is within the interval of $[0.5 \widetilde{\eta}, \widetilde{\eta}]$, where $\widetilde{\eta}$ is the a hyperparameter in our optimization. Because $\Delta w$ is in the same direction of $-g$ (thanks to the post-batch screening), it is guaranteed that a suitable step size always exists. In practice, if the adjustment does not work in few attempts, we simply set the step size small i.e., $\gamma \leftarrow 1.0 \times 10^{-6}$. This however, rarely happens.

Restricting $\eta$ being a very small quantity (e.g., $\widetilde{\eta} = 0.1\%$) is not encouraged. This is similar to the concept of learning rate in SGD. A very small $\eta$ certainly ensures the convergency but $\Delta w$ does not bring a substantial loss reduction. As a second-order method, $E_Q$ is able to approximate $f$ much better than SGD. Therefore, our algorithm is not sensitive to a bigger $\eta$. Unlike learning rate however, $\eta$ is also a relative measure with respect to the actual loss reduction. It is not necessary to frequently adjust this parameter during the optimization.

\paragraph{Discussion}
At first sight, our method could appear similar to trust region method~\cite{sorensen1982newton}, and Taylor ratio $\eta$ resembles the radius of a trust region (normally denoted by $\rho$). A closer look should clarify this confusion: $\rho$ measures the consistence between the reduction of actual loss function and the reduction of $E_Q$, while $\eta$ is more straightforward, indicating the percentage of Taylor error in the loss reduction. More importantly, $\eta$ is used in our method to further adjust the step length. This is fundamentally different from trust region methods, which determine $\rho$ beforehand and stop iteration when $\|\Delta w \|> \rho$. We note that doing so makes choosing $\rho$ troublesome. A bigger $\rho$ leads to big Taylor error, and smaller $\rho$ terminates the iteration too early even the local curvature remains sound and positive. On the other hand, our method allows the algorithm to devote necessary efforts to compute a good search direction locally while avoiding computations for problematic curvatures from second-order CSDD. The step size is adjusted by measuring the ratio between the reduction of $f$ and $E_Q$. This is a noticeable difference from line search methods, which focus mostly on the loss reduction. Finally, our method does not need pesky parameter tweaking. The only hyperparameter is $\widetilde{\eta}$, which is set as $0.05$ in most experiments.

The \textbf{convergency} analysis of Newton-like method is extensively available in numerical computation textbooks e.g.,~\cite{nocedal2006numerical,argyros2008convergence}. The convergency of stochastic Newton method is more involved, yet also well-studied in recent contributions~\cite{kovalev2019stochastic,roosta1601sub,roosta2016sub}. \emph{Our method is globally convergent and has strong local convergency assuming the Hessian is Lipschitz smooth}. We refer the reader to the aforementioned literature for a detailed convergency analysis and move to the experimental study in the next section.

\section{Experiment Results}\label{sec:experiment}
\begin{figure*}[th!]
    \centering
    \includegraphics[width=0.999\linewidth]{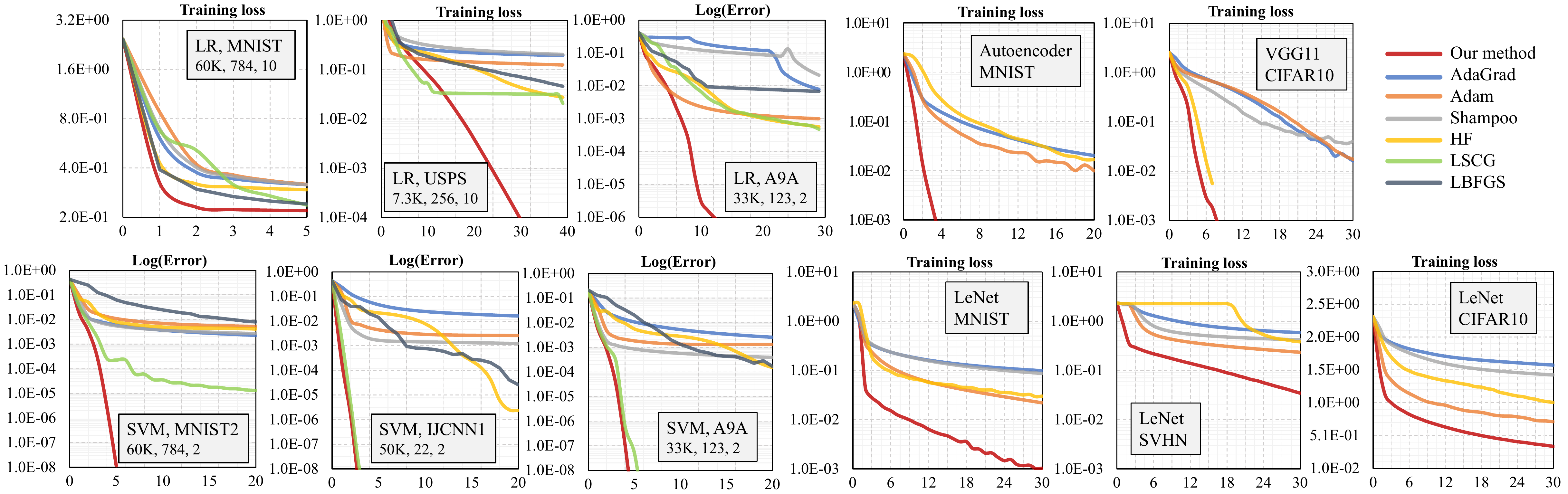}
    \caption{Convergency curves for various deep learning tasks including LR, SVM, autoencoder, and DNNs. The specification of the dataset in LR/SVM tests is also given in the corresponding plot. For instance, \emph{60K, 784, 10} means the total number of training data is 60K. Each date entry has a $784$-dimension feature and will be mapped to $10$ classes/labels.}\label{fig:all}
  \end{figure*}

We implemented our method using \texttt{CuPy} on a desktop PC equipped with an \texttt{Intel} \texttt{i7-6950X} CPU and a \texttt{nVidia} \texttt{1080Ti} GPU. Some more demanding experiments run on a \texttt{2080Ti} GPU. Our method has been tested on various deep learning tasks and compared with multiple well-known optimization algorithms including Adam~\cite{kingma2014adam}, AdaGrad~\cite{duchi2011adaptive}, Shampoo~\cite{gupta2018shampoo}, HF~\cite{martens2010deep}, and LBFGS~\cite{liu1989limited}. The results are reported in Fig.~\ref{fig:all}. In short, \emph{our method outperforms all the competing methods in all the experiments by a noticeable margin. A strong quadratic convergency is observed in the tests}. We refer the reader to the supplementary document for additional performance analysis. The source code and an illustrative video are also accompanied.

\vspace{-5 pt}
\paragraph{Performance comparison between CSFD and AD}
Before plunging into deep learning tasks, we want to confirm that CSFD is indeed a better approach than AD for derivative calculation. To this end, we record the time of calculating $Hp$ for a VGG-19 net~\cite{simonyan2014very} using \texttt{autograd} from \texttt{PyTorch} (\texttt{ver.} \texttt{1.6.0}),  \texttt{GradientTape} from \texttt{TensorFlow} (\texttt{ver.} \texttt{2.3.0}), and CSFD. All the experiments run on the single-thread CPU to avoid interference by different parallelization mechanism. We used a na\"ive CSFD implementation that overloads forward and BP using built-in complex data type from \texttt{NumPy}. CSFD used $58.1~sec.$ on average, which is $14\%$ faster than \texttt{TensorFlow} ($66.4~sec.$) and $27\%$ faster than \texttt{PyTorch} ($73.4~sec.$). It is possible to further accelerate CSFD calculation as in~\cite{luo2019accelerated}. Second-order derivative computation with existing AD routines is not yet feasible and potentially unstable.

\vspace{-5 pt}
\paragraph{Regression \& SVMs}
We compare different training methods for multi-class logistic regression (LR) and support-vector machines (SVMs) using datasets from LIBSVM~\cite{chang2011libsvm}.
Each method is trained with a full batch or minibatches of sizes $128$ or $1,024$, and the best result is recorded for the comparison. In SVM training, we used Hinge-2 loss.
We note that LSCG is rather close to our method in SVM training. We believe this is because of the good convexity of the Hessian in these experiments. This observation reflects a good adaptivity of our method: if the problem is well-shaped, our algorithm becomes a classic Newton-CG routine. However, LSCG does not yield good results in LR training as the least-square Hessian modification defaces the curvature information, and the Hessian-based computation becomes less effective and downgrades LSCG to be first-order convergent.

\vspace{-5 pt}
\paragraph{Deep Neural Networks}
We also evaluate the training convergency on several classic deep neural networks (DNNs) such as deep autoencoder~\cite{kramer1991nonlinear}, LeNet-5~\cite{lecun1998gradient} and VGG-11~\cite{simonyan2014very}. Some methods like vanilla LBFGS and trust region do not work for stochastic training by default. Hence, they are skipped in the corresponding tests. In DNN training, the size of the minibatch is $128$. Our method does not need learning rate tweaking ($\widetilde{\eta}=0.05$). But for other methods, we tried different learning rates from $1.0\times 10^{-1}$ to $1.0\times 10^{-4}$, and the best result is used in the comparison.

Deep \textbf{autoencoder} is trained on MNIST. The architecture of the encoder is $784\rightarrow$ $1,000\rightarrow$ $500\rightarrow$ $250\rightarrow$ $30$. The iterative SVD procedure in Shampoo did not converge, and the corresponding curve is not available. \textbf{LeNet} is a classic DNN architecture with two sets of convolutional layers and average pooling layers, followed by a flattening layer and three FC layers, activated by sigmoid functions. The training is done on MNIST, SVHN, and CIFAR10 datasets. Our experiment also includes the training of a \textbf{VGG} net (VGG-11). The structure of is net is $64\mathtt{Cov}3\rightarrow$ $\mathtt{MP}2\rightarrow$ $128\mathtt{Cov}3\rightarrow$ $\mathtt{MP}2\rightarrow$ $(2\times256\mathtt{Cov}3)\rightarrow$ $\mathtt{MP}2\rightarrow$ $(2\times512\mathtt{Cov}3)\rightarrow$ $\mathtt{MP}2\rightarrow$ $(2\times512\mathtt{Cov}3)\rightarrow$ $\mathtt{MP}2\rightarrow$ $(2\times 512\mathtt{FC})\rightarrow$ $10\mathtt{FC}\rightarrow$ $\mathtt{Softmax}$.
We used ELU as the activation function, and the batch normalization is applied before the activation.

\section{Conclusion}
In this paper, we propose an effective Newton Krylov algorithm of second-order optimization for deep learning. This algorithm is enabled by a novel implementation of differentiation calculations, namely complex-step finite difference or CSFD. CSFD is essentially a complex-overloaded finite difference procedure, and can be conveniently implemented within existing deep learning frameworks. This convenience leads to a better performance and high-order generalization. In this paper, we use CSFD to compute first- and second-order directional derivatives as an initial proof-of-concept study. Indeed, if properly implemented, CSFD has the potential to largely, if not completely replace existing AD methods. Our contribution goes beyond the introduction of numerical differentiation. With the assistance of CSDD, we bring several important improvements over existing stochastic Newton methods. Guided by both global (e.g., gradient) and local (e.g., curvature) information, our Krylov iteration is keen to noise and concavity, and disturbing computations are largely avoided. We combine the advantages of line search and trust region in the stochastic optimization to ensure each improvement is both effective and substantial.

CSFD and CSDD open a new window for future stochastic optimization. We will investigate its parallel and high-order implementation, and deeply couple carefully-crafted numerical procedures with nonlinear optimizations to empower next-generation deep learning techniques.

\bibliography{ref}

\begin{thebibliography}{54}
\providecommand{\natexlab}[1]{#1}
\providecommand{\url}[1]{\texttt{#1}}
\providecommand{\urlprefix}{URL }
\expandafter\ifx\csname urlstyle\endcsname\relax
  \providecommand{\doi}[1]{doi:\discretionary{}{}{}#1}\else
  \providecommand{\doi}{doi:\discretionary{}{}{}\begingroup
  \urlstyle{rm}\Url}\fi

\bibitem[{Abadi et~al.(2016)Abadi, Barham, Chen, Chen, Davis, Dean, Devin,
  Ghemawat, Irving, Isard et~al.}]{abadi2016tensorflow}
Abadi, M.; Barham, P.; Chen, J.; Chen, Z.; Davis, A.; Dean, J.; Devin, M.;
  Ghemawat, S.; Irving, G.; Isard, M.; et~al. 2016.
\newblock Tensorflow: A system for large-scale machine learning.
\newblock In \emph{12th $\{$USENIX$\}$ Symposium on Operating Systems Design
  and Implementation ($\{$OSDI$\}$ 16)}, 265--283.

\bibitem[{Al~Seyab and Cao(2008)}]{al2008nonlinear}
Al~Seyab, R.; and Cao, Y. 2008.
\newblock Nonlinear system identification for predictive control using
  continuous time recurrent neural networks and automatic differentiation.
\newblock \emph{Journal of Process Control} 18(6): 568--581.

\bibitem[{Amari, Park, and Fukumizu(2000)}]{amari2000adaptive}
Amari, S.-I.; Park, H.; and Fukumizu, K. 2000.
\newblock Adaptive method of realizing natural gradient learning for multilayer
  perceptrons.
\newblock \emph{Neural computation} 12(6): 1399--1409.

\bibitem[{Anil et~al.(2020)Anil, Gupta, Koren, Regan, and
  Singer}]{anil2020second}
Anil, R.; Gupta, V.; Koren, T.; Regan, K.; and Singer, Y. 2020.
\newblock Second Order Optimization Made Practical.
\newblock \emph{arXiv preprint arXiv:2002.09018} .

\bibitem[{Argyros(2008)}]{argyros2008convergence}
Argyros, I.~K. 2008.
\newblock \emph{Convergence and applications of Newton-type iterations}.
\newblock Springer Science \& Business Media.

\bibitem[{Becker, Le~Cun et~al.(1988)}]{becker1988improving}
Becker, S.; Le~Cun, Y.; et~al. 1988.
\newblock Improving the convergence of back-propagation learning with second
  order methods.
\newblock In \emph{Proceedings of the 1988 connectionist models summer school},
  29--37.

\bibitem[{Benson and Shanno(2018)}]{benson2018cubic}
Benson, H.~Y.; and Shanno, D.~F. 2018.
\newblock Cubic regularization in symmetric rank-1 quasi-Newton methods.
\newblock \emph{Mathematical Programming Computation} 10(4): 457--486.

\bibitem[{Betancourt(2018)}]{betancourt2018geometric}
Betancourt, M. 2018.
\newblock A geometric theory of higher-order automatic differentiation.
\newblock \emph{arXiv preprint arXiv:1812.11592} .

\bibitem[{Botev, Ritter, and Barber(2017)}]{botev2017practical}
Botev, A.; Ritter, H.; and Barber, D. 2017.
\newblock Practical gauss-newton optimisation for deep learning.
\newblock In \emph{Proceedings of the 34th International Conference on Machine
  Learning-Volume 70}, 557--565. JMLR. org.

\bibitem[{Bottou(2010)}]{bottou2010large}
Bottou, L. 2010.
\newblock Large-scale machine learning with stochastic gradient descent.
\newblock In \emph{Proceedings of COMPSTAT'2010}, 177--186. Springer.

\bibitem[{Chang and Lin(2011)}]{chang2011libsvm}
Chang, C.-C.; and Lin, C.-J. 2011.
\newblock LIBSVM: A library for support vector machines.
\newblock \emph{ACM transactions on intelligent systems and technology (TIST)}
  2(3): 1--27.

\bibitem[{Chapelle and Erhan(2011)}]{chapelle2011improved}
Chapelle, O.; and Erhan, D. 2011.
\newblock Improved preconditioner for hessian free optimization.
\newblock In \emph{NIPS Workshop on Deep Learning and Unsupervised Feature
  Learning}, volume 201. Sierra Nevada Spain.

\bibitem[{Dozat(2016)}]{dozat2016incorporating}
Dozat, T. 2016.
\newblock Incorporating nesterov momentum into adam .

\bibitem[{Duchi, Hazan, and Singer(2011)}]{duchi2011adaptive}
Duchi, J.; Hazan, E.; and Singer, Y. 2011.
\newblock Adaptive subgradient methods for online learning and stochastic
  optimization.
\newblock \emph{Journal of machine learning research} 12(Jul): 2121--2159.

\bibitem[{Forsythe and Wasow(1960)}]{forsythe1960finite}
Forsythe, G.~E.; and Wasow, W.~R. 1960.
\newblock Finite Difference Methods.
\newblock \emph{Partial Differential} .

\bibitem[{Gupta, Koren, and Singer(2018)}]{gupta2018shampoo}
Gupta, V.; Koren, T.; and Singer, Y. 2018.
\newblock Shampoo: Preconditioned stochastic tensor optimization.
\newblock \emph{arXiv preprint arXiv:1802.09568} .

\bibitem[{IEEE(1985)}]{ieee1985ieee}
IEEE. 1985.
\newblock IEEE standard for binary floating-point arithmetic.
\newblock Institute of Electrical and Electronic Engineers.

\bibitem[{Kingma and Ba(2014)}]{kingma2014adam}
Kingma, D.~P.; and Ba, J. 2014.
\newblock Adam: A method for stochastic optimization.
\newblock \emph{arXiv preprint arXiv:1412.6980} .

\bibitem[{Knoll and Keyes(2004)}]{knoll2004jacobian}
Knoll, D.~A.; and Keyes, D.~E. 2004.
\newblock Jacobian-free Newton--Krylov methods: a survey of approaches and
  applications.
\newblock \emph{Journal of Computational Physics} 193(2): 357--397.

\bibitem[{Kovalev, Mishchenko, and Richt{\'a}rik(2019)}]{kovalev2019stochastic}
Kovalev, D.; Mishchenko, K.; and Richt{\'a}rik, P. 2019.
\newblock Stochastic Newton and Cubic Newton Methods with Simple Local
  Linear-Quadratic Rates.
\newblock \emph{arXiv preprint arXiv:1912.01597} .

\bibitem[{Kramer(1991)}]{kramer1991nonlinear}
Kramer, M.~A. 1991.
\newblock Nonlinear principal component analysis using autoassociative neural
  networks.
\newblock \emph{AIChE journal} 37(2): 233--243.

\bibitem[{Lantoine, Russell, and Dargent(2012)}]{lantoine2012using}
Lantoine, G.; Russell, R.~P.; and Dargent, T. 2012.
\newblock Using multicomplex variables for automatic computation of high-order
  derivatives.
\newblock \emph{ACM Transactions on Mathematical Software (TOMS)} 38(3): 1--21.

\bibitem[{LeCun et~al.(1998)LeCun, Bottou, Bengio, and
  Haffner}]{lecun1998gradient}
LeCun, Y.; Bottou, L.; Bengio, Y.; and Haffner, P. 1998.
\newblock Gradient-based learning applied to document recognition.
\newblock \emph{Proceedings of the IEEE} 86(11): 2278--2324.

\bibitem[{Liu and Nocedal(1989)}]{liu1989limited}
Liu, D.~C.; and Nocedal, J. 1989.
\newblock On the limited memory BFGS method for large scale optimization.
\newblock \emph{Mathematical programming} 45(1-3): 503--528.

\bibitem[{Luo et~al.(2019)Luo, Xu, Shao, Xu, and Yang}]{luo2019accelerated}
Luo, R.; Xu, W.; Shao, T.; Xu, H.; and Yang, Y. 2019.
\newblock Accelerated complex-step finite difference for expedient deformable
  simulation.
\newblock \emph{ACM Transactions on Graphics (TOG)} 38(6): 1--16.

\bibitem[{Lyness(1968)}]{lyness1968differentiation}
Lyness, J. 1968.
\newblock Differentiation formulas for analytic functions.
\newblock \emph{Mathematics of Computation} 352--362.

\bibitem[{Margossian(2019)}]{margossian2019review}
Margossian, C.~C. 2019.
\newblock A review of automatic differentiation and its efficient
  implementation.
\newblock \emph{Wiley Interdisciplinary Reviews: Data Mining and Knowledge
  Discovery} 9(4): e1305.

\bibitem[{Martens(2010)}]{martens2010deep}
Martens, J. 2010.
\newblock Deep learning via hessian-free optimization.
\newblock In \emph{ICML}, volume~27, 735--742.

\bibitem[{Martens and Grosse(2015)}]{martens2015optimizing}
Martens, J.; and Grosse, R. 2015.
\newblock Optimizing neural networks with kronecker-factored approximate
  curvature.
\newblock In \emph{International conference on machine learning}, 2408--2417.

\bibitem[{Martens and Sutskever(2011)}]{martens2011learning}
Martens, J.; and Sutskever, I. 2011.
\newblock Learning recurrent neural networks with hessian-free optimization.
\newblock In \emph{Proceedings of the 28th international conference on machine
  learning (ICML-11)}, 1033--1040. Citeseer.

\bibitem[{Martins, Sturdza, and Alonso(2003)}]{martins2003complex}
Martins, J.~R.; Sturdza, P.; and Alonso, J.~J. 2003.
\newblock The complex-step derivative approximation.
\newblock \emph{ACM Transactions on Mathematical Software (TOMS)} 29(3):
  245--262.

\bibitem[{Mizutani and Dreyfus(2008)}]{mizutani2008second}
Mizutani, E.; and Dreyfus, S.~E. 2008.
\newblock Second-order stagewise backpropagation for Hessian-matrix analyses
  and investigation of negative curvature.
\newblock \emph{Neural Networks} 21(2-3): 193--203.

\bibitem[{Mor{\'e}(1978)}]{more1978levenberg}
Mor{\'e}, J.~J. 1978.
\newblock The Levenberg-Marquardt algorithm: implementation and theory.
\newblock In \emph{Numerical analysis}, 105--116. Springer.

\bibitem[{Nasir(2013)}]{nasir2013new}
Nasir, H. 2013.
\newblock A new class of multicomplex algebra with applications.
\newblock \emph{Mathematical Sciences International Research Journal} 2(2):
  163--168.

\bibitem[{Nocedal and Wright(2006)}]{nocedal2006numerical}
Nocedal, J.; and Wright, S. 2006.
\newblock \emph{Numerical optimization}.
\newblock Springer Science \& Business Media.

\bibitem[{Paszke et~al.(2017)Paszke, Gross, Chintala, Chanan, Yang, DeVito,
  Lin, Desmaison, Antiga, and Lerer}]{paszke2017automatic}
Paszke, A.; Gross, S.; Chintala, S.; Chanan, G.; Yang, E.; DeVito, Z.; Lin, Z.;
  Desmaison, A.; Antiga, L.; and Lerer, A. 2017.
\newblock Automatic differentiation in pytorch .

\bibitem[{Paszke et~al.(2019)Paszke, Gross, Massa, Lerer, Bradbury, Chanan,
  Killeen, Lin, Gimelshein, Antiga et~al.}]{paszke2019pytorch}
Paszke, A.; Gross, S.; Massa, F.; Lerer, A.; Bradbury, J.; Chanan, G.; Killeen,
  T.; Lin, Z.; Gimelshein, N.; Antiga, L.; et~al. 2019.
\newblock PyTorch: An imperative style, high-performance deep learning library.
\newblock In \emph{Advances in Neural Information Processing Systems},
  8024--8035.

\bibitem[{Pearlmutter(1994)}]{pearlmutter1994fast}
Pearlmutter, B.~A. 1994.
\newblock Fast exact multiplication by the Hessian.
\newblock \emph{Neural computation} 6(1): 147--160.

\bibitem[{Roosta-Khorasani and Mahoney(2016{\natexlab{a}})}]{roosta1601sub}
Roosta-Khorasani, F.; and Mahoney, M.~W. 2016{\natexlab{a}}.
\newblock Sub-sampled Newton methods I: globally convergent algorithms.
\newblock \emph{arXiv preprint arXiv:1601.04737} .

\bibitem[{Roosta-Khorasani and Mahoney(2016{\natexlab{b}})}]{roosta2016sub}
Roosta-Khorasani, F.; and Mahoney, M.~W. 2016{\natexlab{b}}.
\newblock Sub-sampled Newton methods II: Local convergence rates.
\newblock \emph{arXiv preprint arXiv:1601.04738} .

\bibitem[{Rumelhart, Hinton, and Williams(1986)}]{rumelhart1986learning}
Rumelhart, D.~E.; Hinton, G.~E.; and Williams, R.~J. 1986.
\newblock Learning representations by back-propagating errors.
\newblock \emph{nature} 323(6088): 533--536.

\bibitem[{Saad(2003)}]{saad2003iterative}
Saad, Y. 2003.
\newblock \emph{Iterative methods for sparse linear systems}, volume~82.
\newblock siam.

\bibitem[{Schaul, Zhang, and LeCun(2013)}]{schaul2013no}
Schaul, T.; Zhang, S.; and LeCun, Y. 2013.
\newblock No more pesky learning rates.
\newblock In \emph{International Conference on Machine Learning}, 343--351.

\bibitem[{Schraudolph(2002)}]{schraudolph2002fast}
Schraudolph, N.~N. 2002.
\newblock Fast curvature matrix-vector products for second-order gradient
  descent.
\newblock \emph{Neural computation} 14(7): 1723--1738.

\bibitem[{Simonyan and Zisserman(2014)}]{simonyan2014very}
Simonyan, K.; and Zisserman, A. 2014.
\newblock Very deep convolutional networks for large-scale image recognition.
\newblock \emph{arXiv preprint arXiv:1409.1556} .

\bibitem[{Song, Liu, and Jiang(2019)}]{song2019inexact}
Song, C.; Liu, J.; and Jiang, Y. 2019.
\newblock Inexact proximal cubic regularized Newton methods for convex
  optimization.
\newblock \emph{arXiv preprint arXiv:1902.02388} .

\bibitem[{Sorensen(1982)}]{sorensen1982newton}
Sorensen, D.~C. 1982.
\newblock Newton’s method with a model trust region modification.
\newblock \emph{SIAM Journal on Numerical Analysis} 19(2): 409--426.

\bibitem[{Sutskever et~al.(2013)Sutskever, Martens, Dahl, and
  Hinton}]{sutskever2013importance}
Sutskever, I.; Martens, J.; Dahl, G.; and Hinton, G. 2013.
\newblock On the importance of initialization and momentum in deep learning.
\newblock In \emph{International conference on machine learning}, 1139--1147.

\bibitem[{Toh and Kojima(2002)}]{toh2002solving}
Toh, K.-C.; and Kojima, M. 2002.
\newblock Solving some large scale semidefinite programs via the conjugate
  residual method.
\newblock \emph{SIAM Journal on Optimization} 12(3): 669--691.

\bibitem[{Ueberhuber(2012)}]{ueberhuber2012numerical}
Ueberhuber, C.~W. 2012.
\newblock \emph{Numerical computation 1: methods, software, and analysis}.
\newblock Springer Science \& Business Media.

\bibitem[{Vinyals and Povey(2012)}]{vinyals2012krylov}
Vinyals, O.; and Povey, D. 2012.
\newblock Krylov subspace descent for deep learning.
\newblock In \emph{Artificial Intelligence and Statistics}, 1261--1268.

\bibitem[{Wolfe(1969)}]{wolfe1969convergence}
Wolfe, P. 1969.
\newblock Convergence conditions for ascent methods.
\newblock \emph{SIAM review} 11(2): 226--235.

\bibitem[{Wolfe(1971)}]{wolfe1971convergence}
Wolfe, P. 1971.
\newblock Convergence conditions for ascent methods. II: Some corrections.
\newblock \emph{SIAM review} 13(2): 185--188.

\bibitem[{Zhang, Chen, and Liu(2018)}]{zhang2018local}
Zhang, H.; Chen, W.; and Liu, T.-Y. 2018.
\newblock On the local Hessian in back-propagation.
\newblock In \emph{Advances in Neural Information Processing Systems},
  6520--6530.

\end{thebibliography}


\begin{thebibliography}{4}
\providecommand{\natexlab}[1]{#1}
\providecommand{\url}[1]{\texttt{#1}}
\providecommand{\urlprefix}{URL }
\expandafter\ifx\csname urlstyle\endcsname\relax
  \providecommand{\doi}[1]{doi:\discretionary{}{}{}#1}\else
  \providecommand{\doi}{doi:\discretionary{}{}{}\begingroup
  \urlstyle{rm}\Url}\fi

\bibitem[{Armijo(1966)}]{armijo1966minimization}
Armijo, L. 1966.
\newblock Minimization of functions having Lipschitz continuous first partial
  derivatives.
\newblock \emph{Pacific Journal of mathematics} 16(1): 1--3.

\bibitem[{LeCun et~al.(1998)LeCun, Bottou, Bengio, and
  Haffner}]{lecun1998gradient}
LeCun, Y.; Bottou, L.; Bengio, Y.; and Haffner, P. 1998.
\newblock Gradient-based learning applied to document recognition.
\newblock \emph{Proceedings of the IEEE} 86(11): 2278--2324.

\bibitem[{Simonyan and Zisserman(2014)}]{simonyan2014very}
Simonyan, K.; and Zisserman, A. 2014.
\newblock Very deep convolutional networks for large-scale image recognition.
\newblock \emph{arXiv preprint arXiv:1409.1556} .

\bibitem[{Wolfe(1969)}]{wolfe1969convergence}
Wolfe, P. 1969.
\newblock Convergence conditions for ascent methods.
\newblock \emph{SIAM review} 11(2): 226--235.

\end{thebibliography}

\end{document}


\maketitle
\noindent This document provides more experimental results and performance analysis of our CSDD-based Newton Krylov solver. An illustrative video is also accompanied in the supplementary file.

\section{CSFD vs. FFD in Network Training}
\begin{figure}[ht!]
    \centering
    \includegraphics[width=0.95\linewidth]{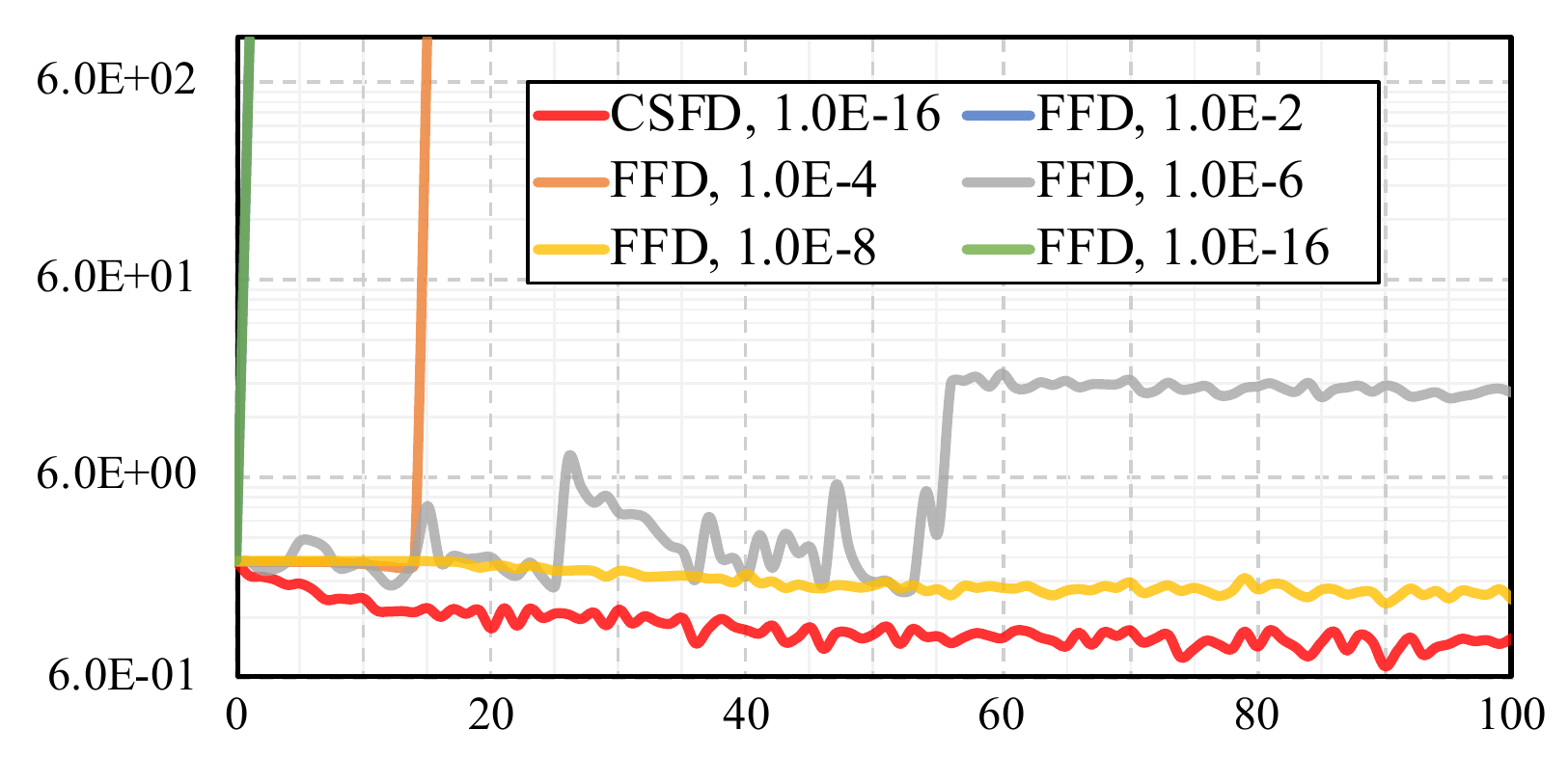}
    \caption{CSFD compared with different FD perturbation settings.}\label{fig:FDVGG}
  \end{figure}

As discussed in the paper, FFD suffers with the subtractive cancellation issue. In our first experiment, we investigate how this issue could impact the actual network training, compared with CSFD. The results are reported in Fig.~\ref{fig:FDVGG}. In this experiment, we use forward finite difference (FFD) with different perturbation sizes to numerically compute the directional derivative. The training network is a VGG-11 net~\cite{simonyan2014very} on CIFAR-10 dataset. The vertical axis in the figure is the per-batch cross entropy loss for the first 100 iterations. As we can see from the plots, if the size of the perturbation is too large e.g., $h = 1.0\times 10^{-2}$ or $h = 1.0\times 10^{-4}$, the approximation error of FFD is significant, and the training diverges quickly (e.g., $h = 1.0\times 10^{-2}$ explodes the training with a single iteration). Decreasing the size of $h$ to $h = 1.0 \times 10^{-6}$ and $h = 1.0\times 10^{-8}$ relieves this issue but does not eradicate it. The loss curve with FFD under $h = 1.0 \times 10^{-8}$ is still oscillating. On the other hand, aggressively reducing $h$ to $h = 1.0 \times 10^{-16}$ crashes the training immediately, because of the subtractive cancellation. We would like to mention that it may be possible that FFD works occasionally with a specific $h$ value. Unfortunately, with deeper and more complicated networks, the chance of finding a working $h$ is slim. On the other hand, CSFD is robust and accurate. Setting $h < \sqrt{\epsilon}$ makes the gradient computation identical to the analytic result.

\section{Ablation Study of Our Method}
Next, we carefully evaluate some key components of our Newton Krylov solver. In this study, we choose to use LeNet-5~\cite{lecun1998gradient} as our primary testbed, and the dataset is MNIST. This is because LeNet-5 is a shallow network, and it overfits MNIST. Therefore, we know \emph{the theoretic global minimum is zero}. In other words, the ground truth is known under this experimental setting. This prior is important for us to better characterize the convergency performance of our method. Otherwise, the actual performance of the algorithm could be veiled by an absolute error measure. For instance, $0.44$ may only appear better than $0.45$ by a nose unless you know the global optimal is at $0.439$. An overfit network training clarifies such potential confusion. Because both the network and dataset are relatively simple, it becomes possible for us to test classic methods (e.g., full deterministic Newton method) even using brute-force Hessian calculation.

\subsection{Early termination}
As an important mechanism, early termination stops the inner Krylov loop as soon as a negative curvature is observed (i.e., $p^\top H p<0$, which is evaluated by $\mathtt{CSDD2}$ routine). Without this condition, our method does not converge -- the loss curve is simply flat.

\subsection{The Taylor ratio $\eta$}
Our method does not rely on sophisticated parameter tuning. The only parameter we need is the range of the Taylor ratio. The Taylor ratio is defined as the ratio between the approximation error of the second-order Taylor expansion and the loss reduction. We exploit this measure during the Krylov iteration to adjust the step size to make sure that $\Delta w$ leads to an $\eta$ within the interval of $\eta \in [0.5\widetilde{\eta}, \widetilde{\eta}]$. It is easy to understand that an over aggressive $\widetilde{\eta}$ downplays the Taylor approximation error making Newton step overshoot; an over conservative $\widetilde{\eta}$ is also harmful by discouraging any ambitious improvements. Thus, the solver easily strands at local minimum.

\begin{figure}[ht!]
    \centering
    \includegraphics[width=0.95\linewidth]{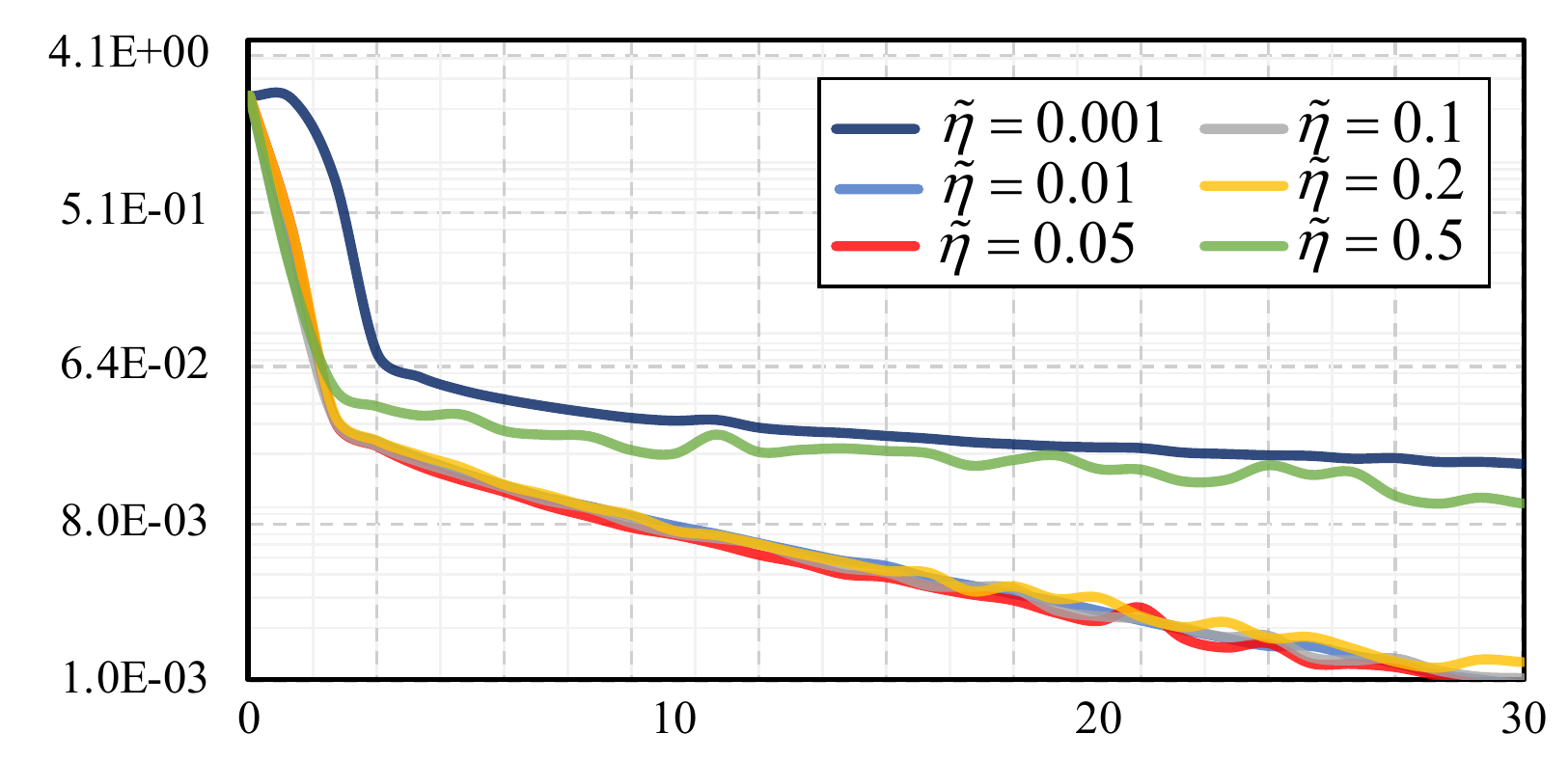}
    \caption{Convergency curves of training LeNet with different $\tilde{\eta}$ settings.}\label{fig:eta}
  \end{figure}

Fortunately, our algorithm is not sensitive to the value of $\widetilde{\eta}$. We would like to emphasize that it does not mean our method converges with any $\widetilde{\eta}$ settings. Yet, it means one does not need to painstakingly figure out a just right $\widetilde{\eta}$. We found that $0.01 < \widetilde{\eta} < 0.2 $ a good option for general training, which yields good convergency. A more thorough study is reported in Fig.~\ref{fig:eta}, where we record the training procedure of our method using different $\widetilde{\eta}$ values, from $0.5$ to $0.001$. It can be seen from the figure that, unless $\widetilde{\eta}$ is extremely set (i.e., $\widetilde{\eta} = 0.001$ or $\widetilde{\eta} = 0.5$), our method has stable performance.


\subsection{Our method vs. backtrack line search}
\begin{figure}[ht!]
    \centering
    \includegraphics[width=0.95\linewidth]{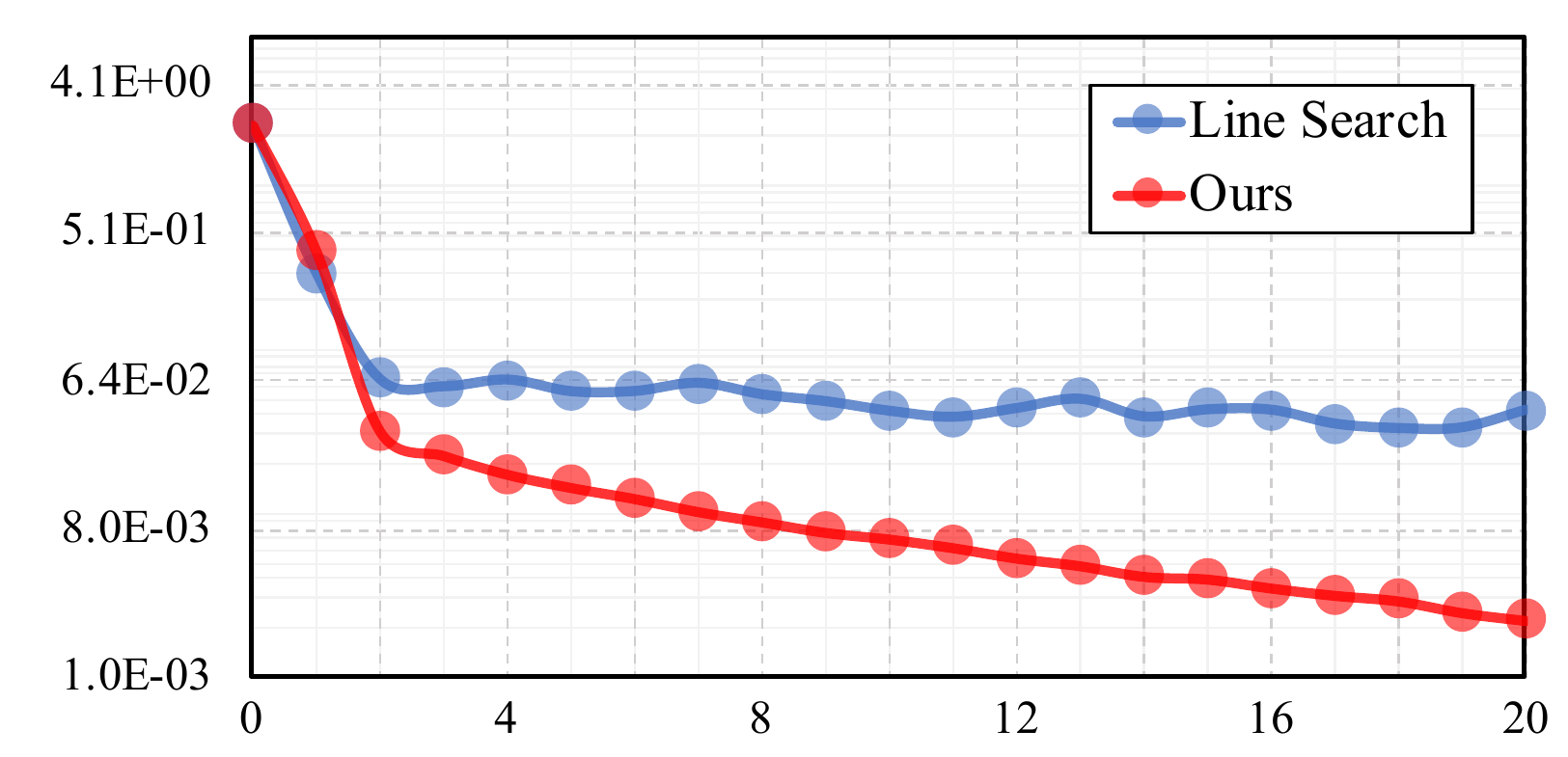}
    \caption{Comparison between our method and backtracking line search.}\label{fig:eta_linesearch}
  \end{figure}

Another question we are curious is how is Taylor ratio based step size adjustment compared with classic line search. To this end, we compare our step sizing method with the backtracking line search, which is often used together with Newton's method. The backtracking procedure starts with a default step size of $\gamma = 1$ (which is same as our method). It reduces the step size by a factor of $0 <\tau < 1$ until the \emph{sufficient decrease} condition or \emph{Armijo} condition~\cite{wolfe1969convergence,armijo1966minimization} is satisfied:
\begin{equation}\label{eq:wolfe}
f(w +\gamma \Delta w )\leq f(w)+\gamma \cdot c \cdot \Delta w ^\top \nabla f(w),
\end{equation}
for some small positive quantity $c$.
We set $c=0.01$ and $\tau = 0.8$ in backtracking line search. The result is compared with our method with $\widetilde{\eta} = 0.05$, and the result is shown in Fig.~\ref{fig:eta_linesearch}. Our method outperforms the line search method. As detailed in Tab.~\ref{tab:linesearch}, our method reduces the training loss an order faster than the line search with only half numbers of attempts of step adjustments.

\begin{table}[ht!]
\centering
\label{tab:eta_linesearch_tries}
\begin{tabular}{c|c|c}
    \hline
    Avg. $\#$ adjustment  & Line search & Our method \\
    \hline
    Epoch 1   & 9.3   &  5.7 ($38.1\%$ less)   \\
    Epoch 10   & 16.7  &  5.1  ($69.8\%$ less) \\
    \hline
\end{tabular}
\caption{Average number of attempts of step size adjustment per iteration at 1st epoch and 10th epoch. }\label{tab:linesearch}
\end{table}

\subsection{Batchsize}
Batch size is always an important parameter in stochastic optimization. In general, a larger batch size gives a more accurate Hessian estimation at a higher computation cost. In this experiment, we report mini-batch-based training loss over the first 200 iterations with different batch sizes, smoothed by moving average.

\begin{figure}[ht!]
    \centering
    \includegraphics[width=0.95\linewidth]{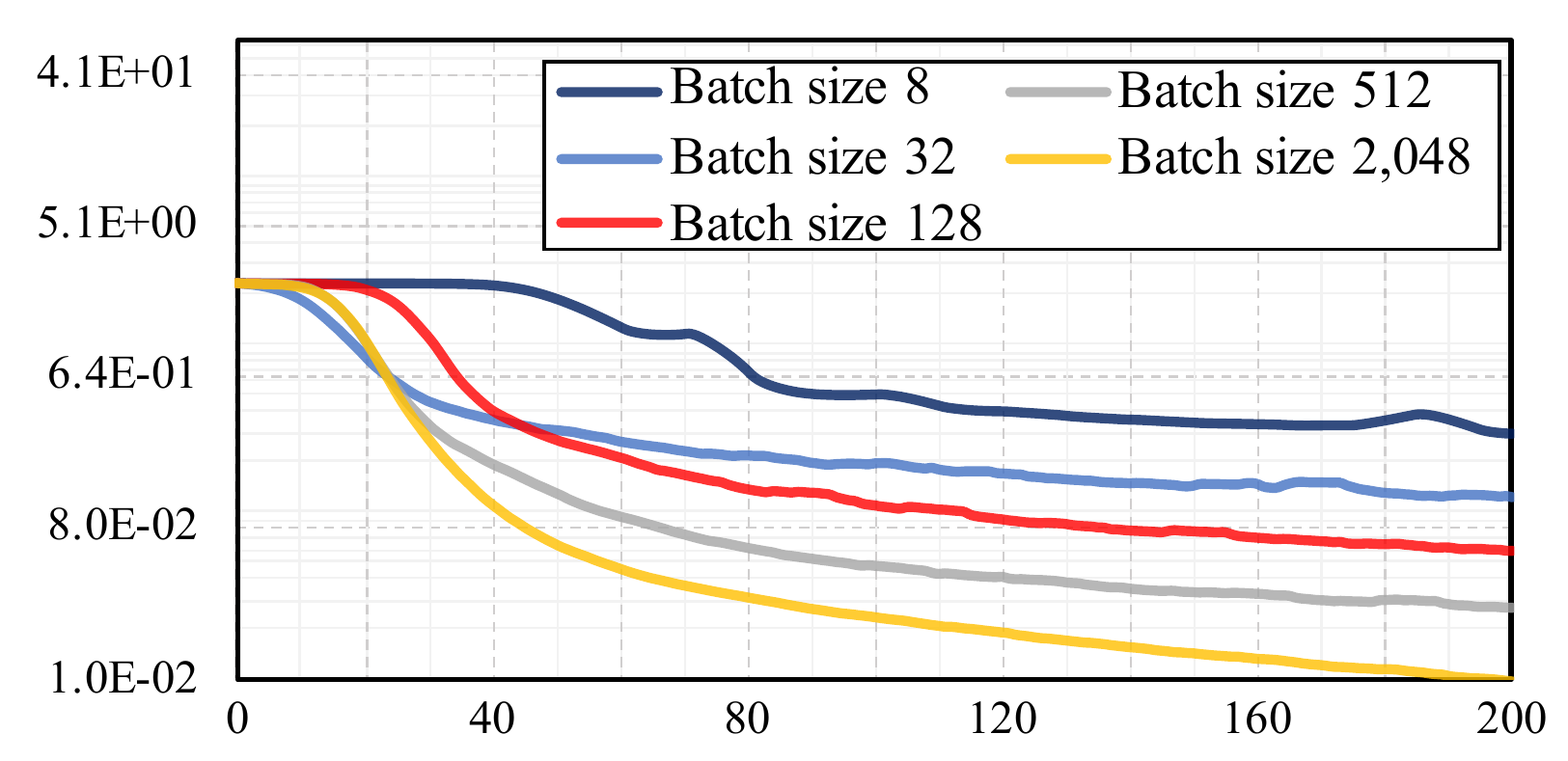}
    \caption{Training LeNet using different batch sizes.}\label{fig:batchsize}
  \end{figure}

\subsection{Activation functions}
An activation function infuses nonlinearity into a neural network and enhances its expressivity. Without proper nonlinear activation and batch normalization (BN), a network degenerates to a simple linear map. In this case, second-order optimization does not have any advantages over first-order methods. This analysis is observed in our experiment.
\begin{figure}[ht!]
    \centering
    \includegraphics[width=0.95\linewidth]{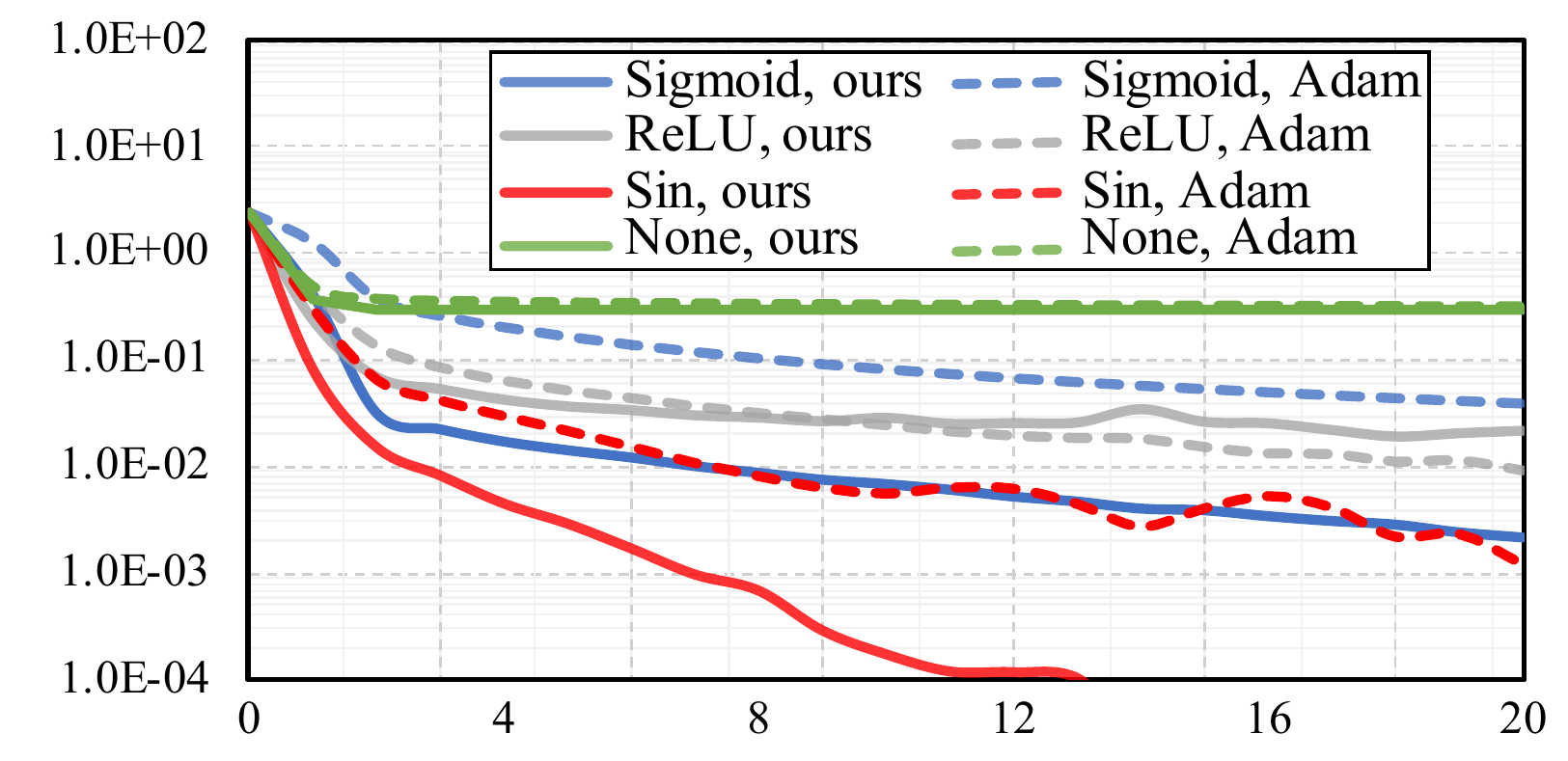}
    \caption{Training LeNet using different activation functions.}\label{fig:activation}
  \end{figure}

Fig.~\ref{fig:activation} reports a test, where we disable BN modules and train the network with several different activation choices. It can be clearly seen from the figure that, without any activation functions (i.e., the green curve labeled as ``None'' in the figure), our method is very close to Adam. During the Krylov iteration, the curvature is always zero, and our method simply becomes a gradient descent procedure. ReLU is a widely-used activation mechanism. Yet, it is also a linear unit: it has a vanished Hessian in most locations. Consequently, the difference between our method and Adam remains marginal (note that BN is not used). Sigmoid gives a different story. It possesses a well-defined second-order around $x=0$, which is better exploited by our method. Therefore, we see a noticeable improvement of using our method over Adam. However, Sigmoid is notorious for its gradient vanishing issue. Without proper treatment, stacking Sigmoid quickly leads the network to be non-differentiable. Finally, we show an interesting test using a trigonometric function $f=\sin(x)$ to activate the network. $f=\sin(x)$ is not only fully smooth with an arbitrary high-order derivative but also periodic. Thus, you are free of gradient vanishing regardless of the range of the input. In this setting, our method significantly outperforms Adam.

This experiment implies that second-order methods are not better than first-order methods by default. In practice, one should choose most suitable optimization procedure based on the dataset pattern and the network architecture. Computing Hessian is expensive. It is always wise to carefully assess the risk and potential benefit before we commit a second-order procedure. For instance, a first-order procedure like Adam should work just fine for training a shallow ReLU-activated net.

\section{Compare with Full Newton-CG}
We also tried to compare our method will full-batch (deterministic) Newton-CG method. Unfortunately, a full-batch Newton-CG does not converge.

\section{Time Statistic}
Our implementation takes about $680~sec.$ on a single epoch for training LeNet on MNIST. The biggest overhead of our algorithm is the calculation of global gradient, which contributes $89.8\%$ of the total computation time. However, computing global gradient can be trivially accelerated and parallelized with more GPUs/CPUs. We implemented our network using \texttt{CuPy} by na\"ive complex overloading, as the complex data type is still under developing on modern deep learning frameworks like~\texttt{PyTorch}. We believe with additional code-level optimizations, CSFD and CSDD have the potential to rebrand many existing deep learning techniques.

\bibliography{supplementbib}